\RequirePackage{fix-cm}

\documentclass[twocolumn]{svjour3}          

\smartqed  

\usepackage{epsfig}
\usepackage{booktabs}
\usepackage{graphicx}
\usepackage{array}
\usepackage{rotating}
\usepackage{pifont}
\usepackage{color}
\usepackage{tabularew}
\usepackage{subfigure}

\journalname{-}

\emergencystretch=\maxdimen
\begin{document}

\title{Robotic frameworks, architectures and middleware comparison}

\author{Tsardoulias, E., Mitkas, A.P.}

\institute{ Emmanouil Tsardoulias \at
              ITI - Information Technologies Institute, CERTH - Centre for Research and Technology Hellas, Thermi 57001, Greece \\
              Tel.: +30 2310 99 6287\\
              \email{etsardou@iti.gr \\
            Pericles Mitkas \at
              Aristotle University of Thessaloniki, Department of Electrical and Computer Engineering, 54124 Thessaloniki, Greece \\
              Tel:	+30 2310 99 6390\\
              \email{mitkas@eng.auth.gr}
              } 
}

\date{Received: date / Accepted: date}

\newcolumntype{P}[2]{%
  >{\begin{turn}{#1}\begin{minipage}{#2}\small\raggedright\hspace{0pt}}c%
  <{\end{minipage}\end{turn}}%
}
\renewcommand\tabcolsep{6pt}
\renewcommand\arraystretch{1.5}

\newcommand{\ty}{\color[rgb]{0,0.7,0} \ding{51}}
\newcommand{\tn}{\color[rgb]{1,0,0} \ding{55}}
\newcommand{\ta}{\color[rgb]{0,0,1} $\sim$}
\newcommand{\tu}{\color[rgb]{0.8,0,0.8} ?}

\maketitle

\begin{abstract}

Nowadays, the construction of a complex robotic system requires a high level of 
specialization in a large number of diverse scientific areas. 
It is reasonable that a single researcher cannot create from scratch the 
entirety of this system, as it is impossible for him to have the necessary 
skills in the necessary fields. 
This obstacle is being surpassed with the existent robotic frameworks. 
This paper tries to give an extensive review of the most famous robotic 
frameworks and middleware, as well as to provide the means to effortlessly 
compare them. 
Additionally, we try to investigate the differences between the definitions of 
a robotic framework, a robotic middleware and a robotic architecture.  


\end{abstract}

\renewcommand{\abstractname}{Acknowledgements}
\begin{abstract}
Parts of this work have been supported by the FP7 Collaborative Project RAPP (Grant Agreement No 610947), funded by the European Commission
\end{abstract}

\section{Introduction}
\label{intro}

Robots are mechanical or virtual agents that are able to perform tasks by 
collecting various types of data and interacting with the environment through 
their effectors. 
It is obvious that since robots are machines (in a wider sense), they are 
incapable of human-like intellectual capabilities such as thinking, taking 
initiatives or improvise. 
On the contrast, robot capabilities are limited to their programmers' software 
that in many cases consists of a large variety of modules from kinematic models 
or PID controllers to high level functionalities, such as navigation strategies 
or vision based object recognition. 
It is apparent that programming a complex robot from scratch is an unfeasible 
task for two reasons. 
First of all a team comprised of scientists specialized in many different 
areas is needed and second, great effort must be applied in joining the 
separate software modules for the whole robotic system to work fluently. 
For that reason, a large variety of robotic frameworks, middleware and 
architecture proposals exist, aiming at not reinventing the wheel, but providing 
a solid basis for any robotics developer to build on and perform in an 
effective way their experiments.

\section{Nomenclature}

A Robotic framework is a collection of software tools, libraries and conventions, 
aiming at simplifying the task of developing software for a complex robotic device. 
In most of the cases, the use of a robotic framework dictates the general 
architectural principles of the developed software (for example if it is 
centralized, real-time etc.). 
It is true that tasks considered trivial by humans are extremely hard to be 
solved in a generic way from a robotic device, as the environmental conditions 
are altered in a dynamic fashion. 
So, in order to create genuinely intelligent robotic software it is essential 
to use a tool like a robotic framework that allows for rapid robotic development, 
as it provides a priori the necessary modules of a robotic system, such as 
safe message passing, motor driving etc.

The Robotic middleware's definition is similar to the one of the Robotic 
framework. 
A descriptive definition of the "robotic middleware" term could be the glue that 
holds together the different modules of a robotic system. 
The most basic task of a robotic middleware is to provide the communications 
infrastructure between the software nodes running in a robotic system. 
The usual case is providing the essential software - hardware interfaces between 
the high level (software) and the low level (hardware) components of the system, 
as these consist of various OS specific drivers that an average robotics 
researcher is impossible to develop. 
It is apparent that the concepts of a robotic framework and a robotic middleware 
are tightly connected and in most of the cases not possible to distinguish. 
A difference that could be noted is that a robotic middleware (if considered 
the glue keeping together the distinct modules) should provide only basic 
functionalities and be invisible to the developer. 
On the contrast, a robotic framework is created to provide the above 
functionality, as well as an API to services or modules already tested by the 
scientific robotic community. In that way it can be assumed that a robotic 
middleware is encapsulated in each robotic framework.

A Robotic architecture differentiates from the robotic framework and robotic
middleware definitions, as it is a more abstract description of how modules in
a robotic system should be interconnected and interact with each other. The
real challenge for a successful robotic architecture is to be defined in such a
way that can be applied to a large number of robotic systems. Obviously this is
a very hard task since robotic systems are characterized by extreme diversity.
For example it is very difficult to define a single architecture that can
operate both with synchronous/asynchronous, single or multiple robot systems.
In conclusion, a robotic architecture should organize a robotics software
system and in the general case, to provide the communication infrastructure
between the different modules (software or hardware).

There are many surveys that present the state-of-the-art in the scientific area
of robotic architectures, middleware and frameworks. For example in \cite{Ref1}
a comparison of open source RDEs (Robotic Development Environments) is being
made, concerning their specifications, platform support, infrastructure,
implementation characteristics and predefined components. Additionally a
comparison exists for each RDE's usability. In \cite{Ref2} eight different
robotic frameworks are compared using three distinct metrics: 'Programming and
communication', 'Robot control patterns' and 'Development and deployment'. Then
the CoRoBa framework is presented in
detail. 

On a similar context, in \cite{Ref3}, \cite{Ref4} and \cite{Ref5} a comparison of
various famous robotic frameworks is being made. On a more theoretical
approach, \cite{Ref6} discusses the consolidation of the two main trends in the
integration frameworks (i.e. the robotic frameworks / middleware and
architectures): the communication layers and the component-based frameworks.
The discussion is being made under the consideration of the bigger picture of a
robotics system, regardless of the actual integration layer, and is presented
through the prism of Rock (the Robot Construction
Kit). 

From the definitions presented it is obvious that the limits of each term are fuzzy
and in many cases overlapping, although the robotic architecture meaning seems
to be more distinct than the other two. For that reason the robotic frameworks
and middleware will be presented in the same chapter (3.2) whilst the robotic
architectures will be described in a separate one
(3.3).

\section{Robotic frameworks and middleware}

The current section contains a state-of-the-art survey of the most famous robotic
frameworks and middleware. The two most relevant frameworks to RAPP are
initially presented (ROS and HOP), then the most wide-spread and the rest are
mentioned in an alphabetic
order.

\subsection{ROS (Robot Operating System)}

The Robot Operating System (ROS) is a framework targeted for writing robot software.
It is comprised of tools, libraries and conventions that aim for complexity
reduction, concerning the procedure of writing complex and robust robotic
behaviours (and generally robotic software). Its creators describe it as
meta-operating system \cite{Ref7}, as it provides standard system operating
services, such as hardware abstraction, low-level device control,
implementation of commonly used functionality and message-passing between
processes and package management. It is fully distributed, as it supports
transparent node execution on heterogeneous robotic devices or computers and is
comprised of three main core
components:

A. Communications infrastructure: The middleware part of ROS is the communications
infrastructure, which is a low-level message passing interface for
intra-process message exchange. Apart from the message passing functionality,
it supports the recording and playback of messages via the rosbag tool, remote
procedure calls, as well as a distributed parameter
system.

B. Robot-specific features: In addition to the core components, ROS is equipped with
various robot-specific tools that speed up the robotic software development.
Some of the most important ones
include:
\begin{itemize}
  \item Standard Message Definitions for Robots
  \item Robot geometry library
  \item Robot description language
  \item Diagnostics
  \item Pose estimation algorithms
  \item Localization modules
  \item Mapping algorithms
  \item Navigation and path creation modules
\end{itemize}

C. Tools: One of ROS strengths is the powerful development toolset. Various
tools are included that provide introspecting, debugging, plotting and
visualizing variables, procedures and even the state of the robotic system.
Using these, the data flow can be easily visualized and debugged, as the
message transmitting system is underlying. The two most famous tools ROS
includes are rviz (for experiment visualization) and rqt (for data
visualization and graphical module incorporation), whereas others such
rosgraph, rxplot etc. provide additional debugging
capabilities.

Finally ROS provides seamless integration with other community accepted robotic libraries
such as the Gazebo 3D simulator, OpenCV for image processing, PCL (Point cloud
library) and MoveIt! for navigation purposes. It must be stated that ROS is not
real time but real-time code can be incorporated with it. Additionally in 2009
the integration of ROS and Orocos RTT (real-time framework - will be described
later) was announced. Conclusively ROS is the state-of-the-art of robotic
frameworks and it tends to be a standard for robotic application
development.

This fact is supported by a grate number of publications concerning ROS. In \cite{Ref6}
ROS is investigated as the tool to traverse from robotic components to whole
systems. Additionally, the distributed characteristic of ROS, as well as some
of its ports to the JavaScript language \cite{Ref9}, allows it to be the link
between intelligent environments and the "internet of things" \cite{Ref10}.
Elkady, Joy and Sobh in \cite{Ref11} use ROS as a plug and play middleware for
sensory modules, actuator platforms and task descriptions in robotic
manipulation platforms, whereas in \cite{Ref12} it is used as basis for the
creation of another framework (CRAM - Cognitive Robot Abstract Machine) for
everyday manipulation in human
environments.

\subsection{Hop}

The Hop system is a toolset for programming the Web of Things \cite{Hop1}. Hop
is typically used to coordinate home automation and robotic environments for
assisted living. Environments consist in the aggregation of communicating
objects (sensors and active components such as robots) which are discovered,
identified, and linked to client/server Hop software modules distributed among
components. Hop orchestrates data and commands transfer among objects, to and
from web services and user interface
components.

Hop is a multi-tier programming environment, built on top of web protocols and languages
(http, web sockets, html, JavaScript). Hop servers run indifferently on PC or
smaller devices, whereas Hop clients run within the JavaScript environment of
web browsers \cite{Hop2}, on PCs and wireless devices. HOP servers may also act
as clients of other HOP servers
\cite{Hop3}.

Hop supports software extensions, enabling the integration of additional objects and
system libraries (for example, Hop provides bindings to control and command
Phidget sensors and actuators, widely used in robotic prototypes). This
approach is very powerful as it allows for arbitrary extensions, at the cost of
some specific integration work to support additional third party
libraries. 

Another integration model is being implemented, promoting the use of JavaScript as the
default programming language for Hop (replacing an ad hoc language, based on
Scheme) and the generic integration of third party software frameworks (such as
ROS). This new approach is expected to dramatically decrease the cost of
integrating new components, such as hardware devices or software
libraries.

HOP applications are called weblets. Applications are written either in the HOP
language or using javascript (Javascript execution is native within web
browsers hosting HOP user interface clients, and a compiler is being prototyped
to also support JavaScript in a HOP server environment). The HOP environment is
extensible; standard libraries provide access to operating system services
(file system, process management, peripherals, network services including
multicast discovery services), whereas additional libraries (such as drivers to
control specific hardware, or  motion control libraries) may be linked to the
environment if needed and made available to application developers through
additional HOP
API.

HOP is designed to allow for the simple specification of distributed applications.
The language syntax allows to direct functions to run either on the server or
the client side, with automatic serialization of client/server messages using
web
protocols. 

\subsection{Player / Stage / Gazebo}

The Player / Stage / Gazebo project is one of the most famous and "traditional"
robotic frameworks. It was initially distributed in 2001 from the USC Robotics
Lab and from then it is being used in various robotic labs and projects
respectfully. As the title suggests, it consists of three distinct software
packages:

A. Player: Player provides a network interface to a large ensemble of robot and
sensor hardware. Its strength lies in the fact that the client/server model it
provides, allows for developing programs in any programming language (that
supports the TCP/IP protocol) and running them in a distributed fashion on a
computer with a network connection to the respective robot. It supports Linux,
Solaris and BSD operating systems. As stated in Player's website "Player
supports multiple concurrent client connections to devices, creating new
possibilities for distributed and collaborative sensing and
control". 

\begin{figure}[ht]
\begin{center}
\includegraphics [width=0.5\textwidth]{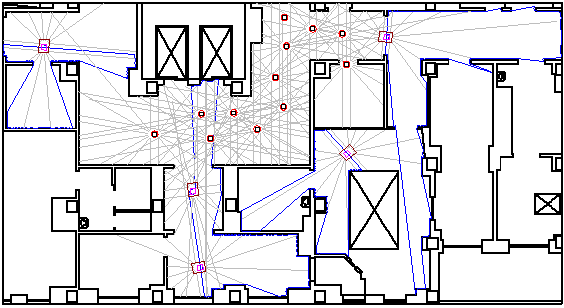}
\caption{Stage 2D simulator} 
\label{fig_2}
\end{center}
\end{figure}

B. Stage : Stage is a 2D (two dimensional) multi-robot simulator. It simulates a variety
of sensors including sonar range finders (SRFs), laser range finders (LRFs),
pan-tilt-zoom cameras and odometry. Stage is created to work seamlessly with
Player, aiming to minimize the software robot controllers needed to perform a
simulation. For that reason many controllers created with Player and tested in
Stage were directly able to work on real robots. A screenshot of Stage is
presented in Fig.
~\ref{fig_2}.

\begin{figure}[ht]
\begin{center}
\includegraphics [width=0.5\textwidth]{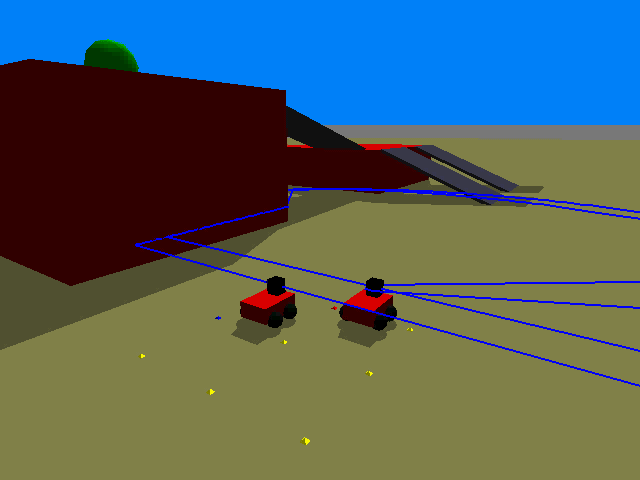}
\caption{Gazebo 3D simulator} 
\label{fig_3}
\end{center}
\end{figure}

C. Gazebo : Gazebo is the third and most recently developed part of the Player/Stage/Gazebo
project and is a 3D (three dimensional) physics multi robot simulator for
outdoor environments. It provides realistic sensor simulation and supports four
different physics engines: ODE (Open Dynamics Engine), Bullet, DART (Dynamic
Animation and Robotics Toolkit from Georgia Tech) and Simbody from Standford
University. Gazebo, except for its native interface, presents a standard Player
interface. In that way the controllers written for the Stage 2D simulator can
be used (in most of the cases) with Gazebo with minimum modifications. A
screenshot of Gazebo simulator is presented in Fig.
~\ref{fig_3}.

Player is referenced in a vast number of publications. In \cite{Ref13} its architecture is
explained and the authors consider it as a "practical" robot programming
framework. Its distributed character is also appraised in many works such as in
\cite{Ref14}, where the massive multi-robot simulating capabilities are
described and in \cite{Ref15} and \cite{Ref16} where the distributed control
services of Player/Stage are presented. Of course, Gazebo has drawn a lot of
attention as well \cite{Ref17} and has been lately supported officially by ROS.
Finally Player/Stage/Gazebo naturally participates in various robotic framework
surveys
\cite{Ref18}. 

\subsection{MSRS (Microsoft Robotics Studio)}

\begin{figure}[ht]
\begin{center}
\includegraphics [width=0.5\textwidth]{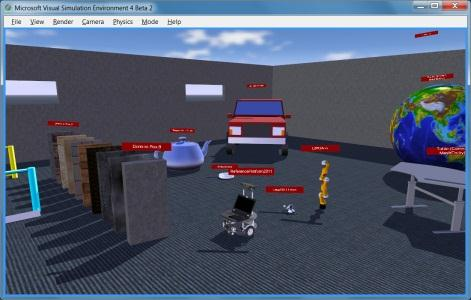}
\caption{3D simulation in MSRS} 
\label{fig_4}
\end{center}
\end{figure}

Microsoft Robotics Studio is a Windows – based environment for robot control and simulation.
It is based on a .NET-based concurrent library implementation for managing
asynchronous parallel tasks, CCR (Concurrency and Coordination Runtime). MRS
implementation involves message-passing and a lightweight services-oriented
runtime called DSS (Decentralized Software Services), which coordinates several
services to orchestrate more complex behaviours. It also provides a 3D
simulation for physics-based virtual environments Fig.
~\ref{fig_4}.

MSRS requires C\# as controller programming language and is not open source. Microsoft
states that MSRS supports a number of robotic devices that can be controlled
either by installing MSRS in them (in an embedded PC running Windows), or
remotely via wi-fi or bluetooth. In \cite{Ref19} the full description and
documentation of MSRS exists, whereas in \cite{Ref20} a technical introduction
in MSRS is attempted. Additionally, in \cite{Ref21}, Microsoft Robotics Studio
is investigated as a mechanism for service – oriented robotics. Finally
\cite{Ref22} compares the Player / Stage / Gazebo framework to MSRS in two
different ways: by examining the documented features of each and by examining
the usability experience gained from implementing two distinct robotic tasks
(wandering and foraging) in a simulated environment. The conclusion reached was
that both frameworks are very capable RDEs (robot developing environments),
though MSRS wins in installation ease whilst Player / Stage / Gazebo is
superior on an architectural
basis.

\subsection{ARIA (Adaptive Robot-Mediated Intervention Architecture)}

ARIA is basically a C++ library that provides various tools, so it can be denoted as
a "pure" robotics framework. These tools include the software input / output
(I/O) integration with custom hardware devices (digital, analog or serial) and,
as its creators state, it supports all MobileRobots / ActivMedia robot
accessories. Additionally, a core part of ARIA is ArNetworking, the tool that
enables the distributed nature of the framework. Specifically ArNetworking
implements an extensible infrastructure for easy remote network operations,
user interfaces and other networked services. Additionally a variety of useful
tools is provided, such as sound effect playback, speech synthesis and
recognition, mathematical functions, cross-platform thread and socket
implementations etc. ARIA has been used for inclusion purposes \cite{Ref23},
where a "children with autism" – specific implementation was
created.

\subsection{ASEBA}

ASEBA is a robotic framework that "provides a set of tools which allow novices to
program robots easily and efficiently". The difference of ASEBA from the
previously described frameworks is that it is an event-based architecture for
real-time distributed control of mobile robots. ASEBA is using a lightweight
virtual machine as core and targets integrated multi-processor robots or groups
of single-processor units, real or simulated. Its strength is that is provides
access to microcontrollers from high level languages, as it integrates with
D-Bus and
ROS.

The real-time and event-based character of ASEBA is referred by various
publications, such as \cite{Ref24}, where is described as a modular
architecture for event-based control of complex robotic systems. In
\cite{Ref25} ASEBA is used in a "games and computer science" context, where an
open-source multiplayer introduction to mobile robots programming is presented.
Finally in \cite{Ref26}, the D-Bus integration to ASEBA is described, that
converts the low-level event architecture into a robotics
middleware.

\subsection{Carmen (Robot Navigation Toolkit)}

Carmen is an open source collection of software for mobile robot control created from the
Carnegie Mellon University. It is constructed in a modular way and provides
basic navigation primitives including base and sensor control, logging,
obstacle avoidance, localization, mapping and path planning. Carmen uses the
inter-process communication platform IPC (InterProcess Communication
facilities) and supports process monitoring. In addition it provides robot
hardware support for different platforms, some of which are iRobot's ATRV and
B21R, ActivMedia's Pioneer I and II, Nomadic Technology's Scout and XR4000, as
well as OrcBoard and Segway. Except for robotic devices, Carmen provides API
functions for robotic sensors such as the Sick LMS laser range finder, GPS
devices using the NMEA protocol, sonars and Hokuyo's IR sensors. Carmen
supports the Linux operating system and the controllers are programmed in the
C++ language. It must be stated that it is not control or real-time
oriented.

At a higher level, it provides a two dimensional robot and sensor simulator. Similarly
to ROS, a centralized parameter server exists, as well as message logging and
playback functionalities. The Carmen framework is written in C but it also
provides Java support, runs under Linux and is available under GPL. In
\cite{Ref27} Carmen is used to showcase a Monte Carlo method for mobile robot
localization, whereas in \cite{Ref28} is presented as a robotics framework in
the context of robot
mapping.

\subsection{CLARAty (Coupled – Layer Architecture for Robotic Autonomy)}

CLARAty was designed and implemented by the Jet Propulsion Laboratory (JPL) of
California Institute of Technology that cooperates with NASA. The developers of
CLARAty describe it as a "reusable robotic software framework". In fact it is a
generic object-oriented framework used for the integration of new algorithms in
many different scientific areas of robotics: motion control, vision,
manipulation, locomotion, navigation, localization, planning and execution.
Additionally it can be used in a number of heterogeneous robots with different
mechanisms and hardware control architectures. From its description and
capabilities, it can be inferred that deviates from the classical "framework"
definition and approaches the one of "robotic architecture". Though, since it
provides a variety of robotic algorithms it is more suitable to be categorized
under the framework / middleware definition. It supports Unix operating systems
and the controllers are written in C++. CLARAty claims to be open-source,
though its development seems to be discontinued, as it was impossible to find a
download webpage or information about development employing
it. 

In \cite{Ref29} CLARAty is investigated under the reusable robotics software
prism. There is a variety of publications about the same subject: In
\cite{Ref30} and \cite{Ref31} Nesnas et al describe CLARAty as a means to
develop interoperable robotic software, whereas in \cite{Ref32} the hardware
heterogeneity is specifically mentioned. In a more general manner, in
\cite{Ref33} a survey is performed concerning CLARATys capability of unifying
the robotic control software mechanisms. Finally \cite{Ref34} and \cite{Ref35}
address its employment under the "autonomous robotics"
perspective.

\subsection{CoolBOT}
	
CoolBOT introduces an interesting parallelism to the robotics software definition. The
main idea is that a software component should be like an electronic component
in a chip. The advantages of an electronic component are that it has a very
clear functionality and a well-established external interface. If this concept
is extended to a robotic system, it could be seen as the integration of
multiple software components. If so, the system's modularity and the software
reusability are maximized. CoolBOT aims at constructing a programming tool,
allowing to program robotic systems by integrating and composing software
components. 

The three main software component types that CoolBOT describes are \textbf{components},
\textbf{views} and \textbf{probes}. \textbf{Components} are contained in
\textbf{integrations}, which is another name for processes.
\textbf{Integrations} can also contain \textbf{views}, which denote essentially
a one-to-one association of an \textbf{integration} and a \textbf{component}.
Finally \textbf{probes} are software components that wrap non CoolBOT software
components with CoolBOT
systems.

CoolBOT can operate both in Windows and Linux and the controllers are developed
in C++. In \cite{Ref36} CoolBOT is described as "a component-oriented
programming framework for robotics", whereas in \cite{Ref37} as a distributed
component-based programming framework for
robotics.

\subsection{ERSP (Evolution Robotics Software Platform)} 

ERSP is a development platform for creating robotic products, meaning that is mostly
oriented in commercial robotic
systems
and not in hobbyist constructions \cite{Ref38}. It provides critical
infrastructure, core capabilities and tools that enable developers to build
robotic applications quickly and easily. It provides an ensemble of robotic
oriented algorithms: ERSP Vision is used for recognizing images and objects in
real – world scenarios and vSLAM help a robotic agent to move autonomously and
with safety. ERSP Navigation incorporate obstacle avoidance, collision
detection, bump detection and cliff detection. Additionally a set of APIs is
included aiming at incorporating the above functionalities in robotic
applications written by developers. The ERSP architecture aims at providing the
developers with standards for structuring their application with a modular way,
as well as tools for combining software and hardware modules seamlessly.
Finally ERSP runs both on Linux and Windows and it is a commercial product. Its
distribution was discontinued since Evolution (the development company) was
acquired from
iRobot. 

\subsection{iRobot Aware ®}

iRobot Aware ® is the robotics framework developed by the iRobot company and is used
in all of its robotic products. Of course Aware is not open source and
information about the operating systems or programming languages it supports
are not available. The latest version of Aware is 2.0 and as iRobot states, "it
utilizes proven industry languages and open source technologies, providing a
flexible and open architecture for third-party development". It provides
advanced robot development tools such as live data viewers, loggers, web-based
data management and other debugging
tools.

\subsection{Marie (Mobile and Autonomous Robotics Integration)}

Marie is a robotic framework / design tool for mobile and autonomous robot applications. It was designed in such a way that can integrate multiple
heterogeneous software elements. It is based on a distributed model, fact that
enables the execution of applications in a group of systems (robots or
computers). Its main goals / motivations
are:

\begin{itemize}
\item To increase the reusability of robotic software components, APIs and even
  frameworks like Player, CARMEN etc., and to support distributed computing in
  heterogeneous platforms.
\item To boost the development process by adopting a rapid-prototyping approach.
\item To allow concurrent use of different communication means (protocols,
  mechanisms, standards).
\item To accelerate user defined development with well-defined layers,
  interfaces, frameworks and plugins .
\item To support multiple sets of concepts and abstractions.
\end{itemize}

MARIE adopts a layered architecture. The three dominant layers are the
\textbf{core layer} which contains tools for low-level OS manipulation, the
\textbf{component layer} which specifies and implements basic frameworks that
enable component incorporation and \textbf{application layer} that contains
tools to create and manage applications using components, aiming at the
creation of a complex robotic system. Having a closer look in the main
architectural blocks of MARIE, there exist four types of
components:

\begin{itemize}
\item Application Adapter (AA) that handles the connection between applications
\item Communication Adapter (CA), enabling for connection of incompatible communication
  mechanisms, as well as for routing functions
\item Application Manager (AM) and Communication Manager (CM). These two
  components are used for managing the different application either locally or
  in a distributed manner
\end{itemize}

Marie is an open source project, though it seems that is no longer maintained.
In \cite{Ref39}, \cite{Ref40} and \cite{Ref41} the software design patterns and
software integration problems that can be solved with the use of Marie are investigated.

\subsection{MCA2 (Modular Control Architecture)}

\begin{figure*}
    \centering
    \subfigure[MCA2 module]
    {
        \includegraphics[width=0.30\textwidth]{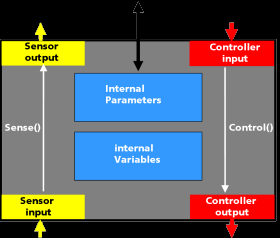}
        \label{fig_mca2_module}
    }
    \subfigure[MCA2 group consisting of modules]
    {
        \includegraphics[width=0.45\textwidth]{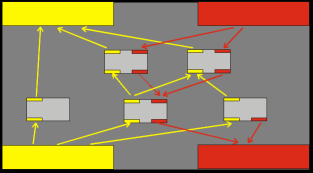}
        \label{fig_mca2_group}
    }
    \caption{MCA2 Architecture}
    \label{fig_mca2}
\end{figure*}

MCA2 is a component-based, real-time capable C/C++ robotic framework for
developing robotic  controllers. Its main parts are called modules (Fig.
~\ref{fig_mca2_module}), which include sensory input and output, controller
input and output, as well as internal parameters and variables. The higher
level consists of groups (Fig. ~\ref{fig_mca2_group}), each of which is a graph
of modules. MCA2 treats groups as modules, so the overall architecture is quite
flexible \cite{Ref42}. MCA2 is network transparent, meaning that groups can be
distributed in a network
environment

The main supporting platform is Linux / RTLinux, but support exists for MCA
OS/X and Win32. MCA2 can also be used in conjunction with visualization /
simulation tools like SimVis3D 4 and other software systems like TinySEP
\cite{Ref43}, a tiny platform for ambient assisted living
\cite{Ref44}.

\subsection{Miro (Middleware for Robotics)}

Miro is a C++, Linux distributed robotic framework that has an object oriented
character, aiming to be used in robotic control. Its is adherent to the common
object request broker architecture (CORBA) standard, fact that enables for
inter-process communication and cross-platform
interoperability. 

Miro was developed in C++ for Linux. Though, since CORBA is programming language independent,
components for Miro can be written in any language or platform that provides
CORBA implementations. The Miro core components were created with the
employment of ACE (Adaptive Communications Environment), which is an object
oriented multi-platform framework for OS-independent interprocess, network and
real time communication. Additionally they use TAO (The ACE ORB) as their
Object Request Broker (ORB), a CORBA implementation designed for high
performance and real time
applications.

Miro architecture consists of three layers:
\begin{itemize}
\item \textbf{Miro Device Layer} is the platform-dependent part of the
  framework and provides abstractions for utilization of robotic sensors and
  actuators.
\item \textbf{Miro Service Layer} contains service abstractions for sensors and
  actuators via CORBA IDL (CORBA Interface Definition Language) and implements
  them as network-transparent modules, allowing for cross-platform systems.
\item \textbf{Miro Class Framework} provides high-level robotic
  functionalities, such as mapping, navigation, path planning, logging and
  visualization.
\end{itemize}

Miro has appeared in a number of publications such as \cite{Ref45} and \cite{Ref46}.

\subsection{MissionLab}

MissionLab was developed by the Mobile Robot Laboratory under the direction of
prof. Ronald Arkin. It takes high-level military-style plans and executes them
with a team of real or simulated robotic vehicles. It supports multi-robot
execution both in simulation and reality. Each vehicle executes its portion of
the mission using reactive control techniques \cite{Ref47}. Programming in
MissionLab occurs in CDL (Configuration Description Language) and CNL
(Configuration Network Language) that after compilation are transformed to C++
code. Also console-like and graphical tools are provided for easy experiment
monitoring. MissionLab uses two servers: HServer (Hardware Server) that
directly controls all the robot hardware and provides a standard interface for
all the robots and sensors, and CBR Server (Case-Based Reasoning Server), which
generates a mission plan based on specifications provided by the user by using
information stored from previous mission plans. Though MissionLab is not widely
used, publications exist that exhibit its capabilities, such as \cite{Ref48}
where a design process for planning in micro-autonomous platforms is described
and \cite{Ref49} where MissionLab is employed in the TAME (Time Varying
Affective Response for Humanoid Robots)
model.

\subsection{MOOS}

MOOS (Mission Oriented Operating Suite) is a C++ cross platform middle ware for
robotics research. It is helpful to think about it as a set of layers:

\begin{itemize}
\item Core MOOS - The Communications Layer: This layer implements a network
  based communications architecture (two libraries and a lightweight
  communications hub called MOOSDB) which enables for building
  inter-communicating applications.
\item Essential MOOS - This consists of applications that deploy CoreMOOS. They
  offer many functionalities such as process control, logging and
  others.
\end{itemize}

Additional tools/applications and libraries are available, with which:

\begin{itemize}
\item a Matlab session can be connected to a set of MOOS enabled processes
\item visually debugging a set of communicating processes is possible
\item replay of logged communications can be performed
\end{itemize}

MOOS has a maritime heritage (its development initiated while the creator was a
postdoc in the Dept. Ocean Engineering at MIT). It is still used for
ocean-bound work \cite{Ref50} and for that reason the full package contains a
few legacy applications which have a maritime
bent:

\begin{itemize}
\item Applications to control NMEA GPS, Orientation and Depth Sensors
\item State - based control of vehicles (pHelm)
\item A pose estimation framework for underwater and surface vehicles (pNav)
\end{itemize}

\subsection{OpenRave} 

OpenRave is oriented in testing, developing and deploying motion planning
algorithms for robots. The main focus is on simulation and analysis of
kinematic and geometric information related to motion planning. For that reason
its use is more oriented to robotic arms and generally in systems that include
many degrees of freedom. It provides command line tools and the run-time core
is easy to be deployed in larger and more complex robotic systems. The most
important technology OpenRAVE provides is a tool called IKFast, the Robot
Kinematics Compiler. This can analytically solve the kinematic equations of any
complex kinematics chain and generate language – specific files, like C++, for
later use. An important target application is industrial robotics automation,
where OpenRave can in theory be easily integrated due to its stand-alone
nature.

OpenRAVE framework was initially created in a thesis and was then expanded
through relevant publications \cite{Ref51}.

\subsection{OpenRDK}

\begin{figure}[ht]
\begin{center}
\includegraphics [width=0.52\textwidth]{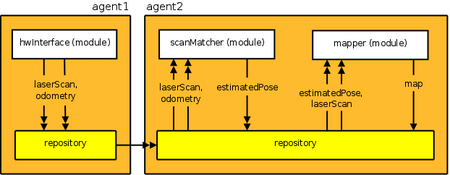}
\caption{Example of two agents in OpenRDK}
\label{fig_openrdk}
\end{center}
\end{figure}

OpenRDK is a modular software framework that intends to accelerate the creation
of complex robotic systems. The main entity is a software process called agent
(Fig ~\ref{fig_openrdk}). A module is a single thread inside the agent process.
Modules can be loaded and started dynamically once the agent process is
running. OpenRDK is distributed, so modules can run in the same or different
machines and communicate using a blackboard-type object, called repository, in
which they publish some of their internal variables called properties. An
extensive description of OpenRDK can be found \cite{Ref52} and
\cite{Ref53}.

\subsection{OPRoS (Open Platform for Robotic Services)}

OPRoS is an open source platform based on components. It provides:
\begin{itemize}
\item An integrated development environment (IDE) for the development of
  robotic software and the monitoring of the robots
\item A framework to manage the distinct software components, including a
  simulator for easier debug and evaluation
\item A server that handles the component repository and specifically the proxy
  services between the different components.
\end{itemize}

A European community for OPRoS exists (since OPRoS is a Corean product), where a
comparison to other robotic frameworks can be found.

OPRoS has been drawing attention since 2008 and there are a number of
publications that include it. In \cite{Ref54}, \cite{Ref55} and \cite{Ref56}
its component-based architecture is being discussed, in \cite{Ref57} attention
is paid in its fault detection capabilities, whereas in \cite{Ref58} an UPnP
single event mechanism for OPRoS is
presented.

\subsection{Orca}

Orca is an open-source C++ framework for developing component-based robotic
systems. It provides the means for defining and developing the building blocks
which can be placed together to form arbitrary complex robotic systems, from
single vehicles to distributed sensor networks. It supports the Linux, Windows
and QNX Neutrino operating systems and its goals are
to 

\begin{itemize}
\item Promote and enable software reuse by standardization of interfaces
\item Provide a high-level and easy to use API, aiming at module reuse
\item Creation of a repository of components
\end{itemize}

In \cite{Ref59} and \cite{Ref60} the component based characteristic of Orca is
presented, whereas in \cite{Ref61} Orca is used as a showcase for the
component-based robotics in general. Finally in \cite{Ref62} the lightweight
characteristics of Orca are described and the advantage of a 'thin' robotic
software frameworks is
discussed.

\subsection{Orocos (Open Robot Control Software)}

\begin{figure}[ht]
\begin{center}
\includegraphics [width=0.5\textwidth]{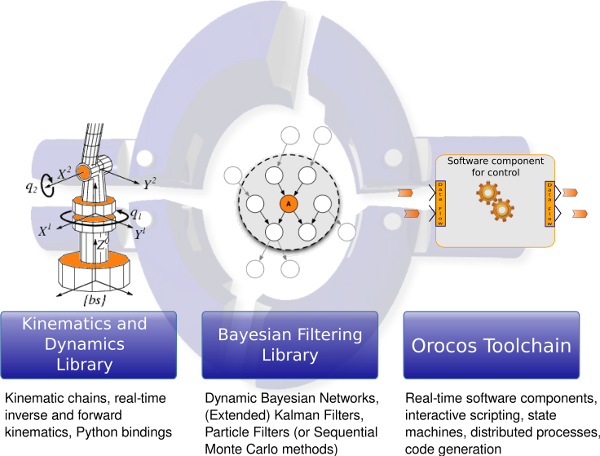}
\caption{The three main modules of Orocos}
\label{fig_orocos}
\end{center}
\end{figure}

Orocos is free software project (basically a collection of portable C++
libraries) oriented in robot control. One of its main characteristics is that
is component based: complex software systems are not constructed at "compile
time" but at "deployment time" or "run time". Additionally it has multi-vendor
support, meaning that components that are built from different vendors can
participate in a more complex system. Finally, its main strength is that it is
free and is focused on \textbf{real time control} of robots and machine tools.
Orocos is comprised of three basic tools (Fig.
~\ref{fig_orocos}): 

\begin{itemize}
\item A Kinematics and Dynamics library which include kinematic chains,
  real-time inverse and forward kinematics
\item A Bayesian Filtering Library which has Dynamic Bayesian Networks,
  (Extended) Kalman filters, particles filters and Sequential Monte Carlo
  methods
\item The Orocos Toolchain that enables for real – time software components,
  interactive scripting, state machines, distributed processes and code
  generation.
\end{itemize}

A detailed description of Orocos can be found in \cite{Ref63}, whereas in
\cite{Ref64} Orocos is used as an architecture for indoor navigation. Finally,
as stated in the ROS description, ROS and Orocos RTT were integrated in 2009,
thus allowing for real-time control in the ROS
environment.

\subsection{RoboFrame}

RoboFrame is a C++ message-oriented middleware. It is designed as a framework
from bottom-up following the concept of inversion-of-control. The platform
specific implementation is wrapped in a thin abstraction layer and currently
the operating systems Linux, BSD Linux, Windows are supported. RoboFrame
provides message exchanging and shared memory communication mechanisms having
in mind two important
parameters:

\begin{itemize}
\item Enabling of message exchange between components running in the same
  thread using references without any overhead commonly present
\item Enabling remote monitoring of the robotic device, taking under
  consideration the specific parameters of the robot-to-base station
  communication difficulties.
\end{itemize}

RoboFrame is described as a modular software framework for lightweight
autonomous robots in \cite{Ref65}. Finally it provides a graphical user
interface that enables for easier robotic software development, though no
information was discovered about if it is open-source or about downloading and
employing the
package.

\subsection{RT middleware (Robotics Technology Middleware) – OpenRTM-aist}

RT middleware is a collection of standards for robots, supporting distributed
execution. The basic unit of this framework is the RT-component which is a
generic class of robotic "objects", such as actuators. RT-middleware uptakes
the task of creating a network graph consisting of RT-components in order to
create a more complex system, aiming at code and consecutively component
reusability. Each RT-component is attached to another RT-component via a
"port". There are many types of ports and only common types of ports can be
connected. Each RT-component is stateful, in the meaning that it can be
perceived as a state
machine. 

OpenRTM-aist is a software platform based on the RT middleware standard. It
implements some real time features and includes a manager that allows easier
manipulation of
RT-Components.

A general description of RT-middleware, including its overall architecture, as
well as the functionalities of RT-Components can be found in the work of Ando,
Noriaki et al \cite{Ref66}, \cite{Ref67},
\cite{Ref68}.

\subsection{Pyro (Python Robotics)}

Pyro stands for Python Robotics and its main goal is to promote the high-level
concept of programming robotic systems. This means that the developer will pay
minimum attention to low-level implementation details, allowing him to easily
concentrate to more interesting advanced robotic topics such as artificial
intelligence, multi-robot planning etc. Its main features
are: 

\begin{itemize}
\item It is open source, so it is available for expansion
\item Designed for research and education use
\item Works on many heterogeneous robotics platforms and simulators
\item Encapsulates a rich module repository, comprising control methods, vision
  (motion tracking, blobs, etc.), learning (neural networks, reinforcement
  learning, self-organizing maps, etc.), evolutionary algorithms, and more.
\end{itemize}

As its name suggests, Pyro is written in Python, which means that the
researcher can interactively experiment with the robot programs. Of course, as
a middleware, it abstracts all of the underlying hardware details, enabling for
easy diverse robot integration. Pyro is often presented as a convenient tool
for teaching robotics \cite{Ref69}\cite{Ref70}, as well as for more
sophisticated use, e.g. artificial intelligence in robotics
\cite{Ref71}\cite{Ref72}. Finally it must be stated that Pyro is used both in
research and in education as a
courseware.

\subsection{ROCI (Remote Objects Control Interface)}

Remote Objects Control Interface (ROCI) is a middleware that provides software
developers the tools to construct a sensing and processing network, in a manner
that can be easily managed. As most of the robotic frameworks, it uptakes the
task of providing the underlying infrastructure for message passing and
generally the whole system inter-communication. Some more specific features
include thread management, peer to peer file search and download, process
sandboxing, remote system management and logging.

As far as architecture is concerned, ROCI is a collection of \textbf{ROCI
modules}, each of which is a process: it takes input, performs computations and
exports an output. An ensemble of ROCI modules is called a \textbf{ROCI task}
which is in fact the conception of a complete functionality. In other words a
ROCI task contains the needed ROCI modules to complete a goal. Finally a
\textbf{ROCI node} is a collection of ROCI tasks and plays a more
organizational than functional
role.

Its distributed manner and its ability to work on a multi-robot team at the
same time are discussed in publications like \cite{Ref73}, \cite{Ref74} and
\cite{Ref75}.

\subsection{RSCA (The Robot Software Communications Architecture)}

RSCA tries to standardize the whole development procedure of robotic
applications. One of its main strengths is that supports real-time
functionalities in conjunction with a communication middleware and a deployment
middleware. The latter manages the components which comprise the while robotic
system, allowing the robotics researcher to install, uninstall, create, start,
stop and tear-down
them.

In designing RSCA, a middleware called SCA from the software defined radio
domain was adopted and extended, since the original SCA lacks the real-time
guarantees and appropriate event services
\cite{Ref76}.

\subsection{ROCK (The Robot Construction Kit)}

ROCK is a software framework that is used for the development of robotic
systems. The whole system is based on the Orocos RTT framework and increases
its functionality by providing ready-to-use drivers and modules. Of course it
makes it possible for the provided ensemble of components to be enriched by
other external algorithms. ROCK was specifically constructed in order to give
solution to the following
issues:

\begin{itemize}

\item Robustness and time-sustainability of robotic systems. It provides error
  detection, reporting and handling mechanisms that allow a system to be
  maintained in an easy way. 
\item Scalability and easy extension. Tools are provided for complex systems
  management, though they are not essential for a robotics researcher to start
  developing. Instead, components can be manipulated with oroGen, the ROCK's
  component manager, in a high-level fashion.
\item Repository of modules and reusability. As any other robotic framework,
  ROCK allows for easy, off-the-shelf employment of diverse algorithms in a
  modular way. Specifically, ROCK's architectural functionality is designed to
  be independent of the framework itself, allowing for module integration with
  other frameworks.

\end{itemize} 

\subsection{SmartSoft}

Smartsoft is a framework that focuses on a component approach of a complex
robotics system and specifically, it treats the proposed communication patterns
as the core of its component model. One important aspect it provides - and
differentiates it from other approaches - is the dynamic, on-line wiring of the
components, meaning that it allows for loose coupling systems that can be
dynamically reconfigured during execution, under specific conditions. Generally
Smartsoft's ambition is to help A) the component developer, B) the application
builder and C) the end user, to create in an easy manner applications by
merging different software components, whose inter-communication is predefined.
Smartsoft is a UNIX framework and the controllers are developed in
C++.

The real – time strengths of SmartSoft are described in \cite{Ref77}. Finally
in \cite{Ref78} SmartSoft is used for developing an application for
sensorimotor
systems.

\subsection{TeamBots}

TeamBots is a collection of Java packages and algorithms for multi-agent
robotics research. It supports prototyping, simulation and execution of
multi-robot control systems \cite{Ref79}. Robot control systems developed in
TeamBots can run in simulation using the TBSim simulation application. One
important aspect is that it provides seamless execution of the robotics
software in real robots using the TBHard robot execution environment, meaning
that the same control systems can run both in simulation and real world. The
fact that TeamBots is based on the Java programming language has its pros and
cons. The obvious advantage is that it extremely portable, as it can operate
under the Windows, Linux, MacOS or any other operating system that supports
Java. On the other hand Java is infamous about its performance compared to C or
C++, due to the fact that the programs developed in Java are executed in a
virtual machine, instead of the real operating
system.

\subsection{Urbi and Urbiscript}

Urbi is not a robotics framework in its strict sense, as it allows to model any
complex system in general. Its component library is developed in C++ and is
called UObject, providing a standardized way to specify and use motors, sensors
and algorithms. The interaction between the different modules of the system is
performed by the urbiscript utilization, a script language by which high level
behaviours can be described. It has similarities to Python or LUA, but
additionally supports embedded parallel and event-driven semantics.
Conclusively urbiscript is a robotics programming language that supports
concurrency and event-based programming, thus can be used in asynchronous
systems.

The goal of Urbi is common to the majority of the robotic frameworks and is to
help making robots compatible, by seamless integration of diverse software
modules. In \cite{Ref80} and \cite{Ref81} the universality of Urbi regarding
its capabilities as a robotic platform is presented, whereas \cite{Ref82}
describes a design of software architecture for an exploration robot based on
Urbi. Urbi now supports ROS, so a developer has a more extended toolset to
produce a robotic application by combining the strengths of both
frameworks.

\subsection{Webots}

Webots is a commercial development environment, developed by Cyberbotics, used
for modelling, programming and simulating mobile robots. It provides
multiple-robot (heterogeneous) simulation in a shared environment that allows
for physical collaboration of robotic agents. Each robot, sensor and generally
any object can be altered, regarding its basic properties such as shape, mass,
friction, color, texture etc. The robot behaviours implemented can be tested in
physically realistic worlds, as a 3D physics engine is employed (ODE - Open
Dynamics Engine). Additionally it has interfaces with Matlab, ROS and Urbi and
can collaborate with CAD software packages such as SolidWorks, AutoCAD, Blender
and others. Webots is available for Windows, Linux and OS/X. In \cite{Ref83}
and \cite{Ref84} the WebotsTM is
presented.

\subsection{YARP (Yet Another Robot Platform)}

The YARP framework incorporates a collection of libraries, protocols and tools,
aiming to clearly decoupled modules. Its architecture promotes a transparent
way of inter-connecting the different modules, in a manner that possible future
alterations (change sensors, actuators) or expansions (add modules) can be
performed with minimum effort. YARP is open-source, supports Windows, Linux,
OS/X and Solaris and the modules are developed in the C++ language. Its main
parts (components)
are:

\begin{itemize}

\item libYARP\_OS - Uptakes the task of cross-OS functionality, as it
  interfaces the applications with the operating system. This library is written
  to be as generic as possible, by using the ACE (Adaptive Communication
  Environment) library, which is portable, fact that enables YARP to be portable
  too.
\item libYARP\_sig - performing common signal processing tasks (visual,
  auditory) in an way that can be easily interfaced with other robotic (and not
  only libraries) such as OpenCV.
\item libYARP\_dev - Uptakes the task of interfacing with hardware robotic
  devices, such as cameras, motor control boards etc.

\end{itemize}

A presentation of YARP can be found in \cite{Ref85}, whereas in \cite{Ref86}
YARP is used in the context or robotic vision applications. Finally in
\cite{Ref87} the modular capabilities of YARP are described. There the software
pieces are described as 'robot genes' and YARP helps in their 'preservation',
meaning the code
reuse. 

\section{Robotic architectures}

The current chapter will briefly introduce some examples of robotic architectures. 

\subsection{AuRA}

The official description of AuRA \cite{Ref126} is followed by the "Principles
and Practice in Review" phrase. AuRA is a hybrid deliberative / reactive
robotic architecture. The high level part is the deliberative component, whilst
the low level part is a reactive behavioural control scheme. An example of a
system that uses AuRA is TAME \cite{Ref127}, a framework for affective robotic
behavior. It must be stated that the MissionLab system itself is a version of
AuRA.

\subsection{BERRA}

Another example of robotics architecture is BERRA \cite{Ref128}. This robot
architecture is again of the hybrid deliberative/reactive behavioural type. The
architectural scheme is divided in three
layers:
\begin{itemize}

\item Deliberative layer that holds the higher-level plans
\item Task execution layer that includes the tasks needed to complete the higher
  level plans
\item Reactive layer that serves whatever reactive low-level events could occur
  while executing the tasks

\end{itemize}
Additionally, the specific architecture is designed to be scalable and flexible,
aiming at minimum effort system reconfiguration.

\subsection{DCA}

DCA (Distributed Control Architecture) is described in \cite{Ref129}. There,
the need for distributed software is addressed, not only in different
executables, but in a network of computers as well. Thus, a distributed
architecture is proposed, that primarily aims at robot control, but of course
can be applied in a broad spectrum of applications. The main design decisions
were based on the following
concepts.

\begin{itemize}

\item Modularity: The programming language used by DCA is modular in nature and
  special attention was paid in constructing the architecture in a way that the
  implementation and incorporation of new modules is facile.
\item Scalability: The DCA programming language supports hierarchical
  implementation of systems. That way a complex system can be scalable if
  designed with a distributed fashion.
\item Efficiency: A separation of real-time event and non-real-time event is
  being made, in order to increase the efficiency of the overall
  system.
\item Flexibility and generality: The type of communication scheme used, allows
  for duplex communication among single or teams of nodes.
\item Theoretical foundation: As the publication authors state "This was
  addressed by using a process algebra adopted from a formal model of
  computation".

\end{itemize}

\subsection{Saphira}

The Saphira architecture is an integrated sensing and control system for
robotics applications. Initially was developed for use on the research robot
Flakey at SRI International and then was formulated in an actual architecture
for development of robotic systems.

The initial Saphira system was divided into two parts. The low-level part was
separated and formulated in another framework, Aria (see subsection 3.5),
maintained by ActivMedia Robotics. As stated, it is a C++ framework oriented in
robot control.

Aria provides a set of communication routines that allow for easy linking with
client programs. This enables researchers to create control systems, without
the need of dealing with low level
details. 
 
On top of the Aria subsystem is a robot control architecture, that addresses
problems like navigation, path planning, object recognition, relying on Aria
for interfacing with the hardware
part. 

\subsection{GenoM}

The Generator of Modules (GenoM) \cite{RefGenom} is a tool to design real-time
software architectures and is oriented to complex embedded systems. The
requirements that GenoM satisfies are:

\begin{itemize}
\item The integration of a wide algorithmic variety that can have different
  real-time constraints and complexities. 
\item A standardized system to incorporate the above functions, in the context
  of time control and data flow.
\item The management of the parallelization, the physical distribution and the
  portability of the functions
\end{itemize}

Finally many review publications exist that investigate the differences between
various robotic architectures. In \cite{Ref130} three architectures are studied
more closely, Saphira, TeamBots and BERRA. Qualities such as portability, ease
of use, software characteristics, programming and run-time efficiency are
evaluated. In order to get a true hands-on evaluation, all the architectures
are implemented on a common hardware robot platform. A simple reference
application is made with each of these systems. All the steps necessary to
achieve this are discussed and compared. Run-time data are also gathered.
Conclusions regarding the results are made, and a sketch for a new architecture
is made based on these
results.

\section{Conclusions}

In this paper a presentation of the most "famous" robotic frameworks,
middleware and architectures is performed. The presented frameworks were of
wide diversity in the operation systems they support, the programming languages
and their orientation (simulation, control, real-time etc.). Additionally, the
survey contained both commercial and open-source software. Here, it must be
stated that the software packages described could incorporate simulators, but
\textbf{pure} simulators were not
reviewed. 

In Table 1 a comparison of the robotics frameworks described in section 3 is
attempted. The metrics used are:

\begin{itemize}

\item a. The operating systems supported by the framework
\item b. Programming languages that can be used to develop applications with
\item c. If it is open source
\item d. If its architecture supports distributed execution
\item e. If it has HW interfaces and drivers
\item f. If it contains already developed robotic algorithms
\item g. If it contains a simulator
\item h. If it is control, or real-time oriented

\end{itemize}

For the metrics c to h the following symbols are used:

\begin{itemize}

\item \ty : Yes
\item \tn : No
\item \ta : Not entirely accurate
\item \tu : Sources not found

\end{itemize}

\begin{table*}[!ht]
\caption{Robotic framework and middleware comparison}

  \begin{tabularew}{c|c|c|c|c|c|c|c|c|c|c}
  \toprule
	\multicolumn{1}{P{90}{2.0cm}}{RFWs} &
    \multicolumn{1}{P{90}{2.0cm}}{OS} &
    \multicolumn{1}{P{90}{2.0cm}@{}}{Programming language} &
    \multicolumn{1}{P{90}{2.0cm}@{}}{Open source}  &
    \multicolumn{1}{P{90}{2.0cm}@{}}{Distributed architecture}  &
    \multicolumn{1}{P{90}{2.0cm}@{}}{HW interfaces and drivers}  &
    \multicolumn{1}{P{90}{2.0cm}@{}}{Robotic algorithms}  &
    \multicolumn{1}{P{90}{2.0cm}@{}}{Simulation}  &
    \multicolumn{1}{P{90}{2.0cm}@{}}{Cοntrol / Real\-time oriented}
    \\
  \midrule
  	ROS & Unix & C++, Python, Lisp & \ty & \ty & \ty & \ty & \ta & \tn \\ \hline
  	
	HOP & Unix, Windows & Scheme, Javascript & \ty & \ty & \ta & \tn & \tn & \tn \\ \hline

	Player/Stage/Gazebo & Linux, Solaris, BSD & C++, Tcl, Java, Python & \ty & \ta & \ty & \ty & \ty & \tn \\ \hline

	MSRS (MRDS) & Windows & C\# & \tn & \ty & \ta & \tn & \ty & \tn \\ \hline
	
	ARIA		 		& Linux, Win & C++, Python, Java & \ty & \tn & \ty & \ty & \tn & \tn \\ \hline
	
	Aseba				& Linux & Aseba & \ty & \ty & \ty & \tn & \ta & \ty \\ \hline
	
	Carmen 				& Linux & C++ & \ty & \ty & \ty & \ty & \ty & \tn \\ \hline
	
	CLARAty 			& Unix & C++ & \ty & \ty & \ty & \ty & \tn & \tn \\ \hline
	
	CoolBOT 			& Linux, Win & C++ & \ty & \ty & \ta & \tn & \tn & \tn \\ \hline
	
	ERSP 				& Linux, Win & \tu & \tn & \ty & \ty & \ty & \tn & \tn \\ \hline
	
	iRobot Aware		& \tu & \tu & \tn & \tu & \ty & \tu & \tn & \tu \\ \hline
	
	Marie				& Linux & C++ & \ty & \ty & \ty & \tn & \tn & \tn \\ \hline
	
	MCA2				& Linux, Win32, OS/X & C, C++ & \ty & \ty & \ty & \tn & \tn & \ty \\ \hline
	
	Miro				& Linux & C++ & \ty & \ty & \ty & \tn & \tn & \tn \\ \hline
	
	MissionLab			& Linux, Fedora & C++ & \ty & \ty & \ty & \ty & \ty & \tn \\ \hline
	
	MOOS				& Windows, Linux, OS/X & C++ & \ty & \ta & \ty & \ty & \tn & \tn \\ \hline
	
	OpenRAVE			& Linux, Win & C++, Python & \ty & \tn & \tn & \ty & \ty & \tn \\ \hline
	
	OpenRDK				& Linux, OS/X & C++ & \ty & \ty & \ty & \tn & \tn & \tn \\ \hline
	
	OPRoS				& Linux, Win & C++ & \ty & \ty & \ty & \ty & \ty & \tn \\ \hline
	
	Orca				& Linux, Win, QNX Neutrino & C++ & \ty & \ty & \ty & \ta & \tn & \tn \\ \hline
	
	Orocos				& Linux, OS/X & C++ & \ty & \ty & \ty & \ty & \tn & \ty \\ \hline
	
	RoboFrame			& Linux, BSD, Win & C++ & \tu & \ty & \ty & \tn & \tn & \tn \\ \hline
	
	RT middleware		& Linux, Win, CORBA platform & C++, Java, Python, Erlang & \ty & \ty & \ty & \tn & \tn & \tn \\ \hline
	
	Pyro				& Linux, Win, OS/X & Python & \ty & \tn & \ty & \ty & \ty & \tn \\ \hline
	
	ROCI				& Win & C\# & \ty & \ty & \tn & \tn & \tn & \tn \\ \hline
	
	RSCA				& \tu & \tu & \tn & \tn & \ty & \tn & \tn & \ty \\ \hline
	
	ROCK				& Linux & C++ & \ty & \tu & \ty & \ty & \tn & \ty \\ \hline
	
	SmartSoft			& Linux & C++ & \ty & \ty & \tn & \tn & \tn & \tn \\ \hline
	
	TeamBots			& Linux, Win & Java & \ty & \tn & \ty & \ty & \ty & \tn \\ \hline
	
	Urbi (language)		& Linux, OS/X, Win & C++ like & \ty & \tn & \ty & \tn & \tn & \tn \\ \hline
	
	Webots				& Win, Linux, OS/X & C, C++, Java, Python, Matlab, Urbi & \tn & \tn & \ty & \tn & \ty & \tn \\ \hline
	
	YARP				& Win, Linux, OS/X & C++ & \ty & \ty & \ty & \tn & \ty & \tn \\
 
  \bottomrule
  \end{tabularew}

\end{table*}

Some conclusions extracted from the review follow.

The first conclusion is that there has been a great effort in creating a large
number of robotic frameworks, fact that indicates that modern robotic
development is extremely hard to perform from scratch. This is supported by the
extreme specialization that currently exists in the robotic scientific area,
forcing the researchers to make use of already developed tools in order to
construct a complex robotic system, or perform a complete test. These
circumstances led the scientific community to create a number of robotic
frameworks that best suited their needs. Of course there are cases of
frameworks that officially support usage in conjunction with others, increasing
their capabilities and potential target
group. 

Currently, the robotic framework with the larger momentum and appeal to the
community is ROS (Robot Operating System). It is open source and extremely
modular with great tools, such as rviz for visualization and tf for
investigation of geometric transforms both in space and time. The scientific
community embraced it, fact supported from the large amount of submitted
algorithms in the ROS wiki that cover a wide range of applications, from
mapping and navigation to motor control. ROS was initially developed in 2007 by
the Standford Artificial Intelligence Laboratory, then in 2008 the development
continued from Willow Garage and in Feb. 2013 ROS transitioned to OSRF, the
Open Source Robotics
Foundation.

An interesting fact is that only a small portion of the presented frameworks is
oriented towards real-time control (Orocos, RSCA, MCA2 etc), something that
shows a trend in modern robotics, meaning that the majority of researchers are
investigating higher-level problems. Additionally the vast majority of the
frameworks have a modular (component-based) architecture in order to increase
code reusability and minimize the integration efforts. Of course, the
modularity can play a crucial role in the following years, as the cloud
computing and cloud robotics begin to
expand.

The number of frameworks that supported Linux or Unix is larger than the ones
that operate in Windows or other operating systems. Also an important fraction
of the ones that supported Windows were commercial, something that reveals the
open-source tradition of Unix systems, opposed to the Windows OSs. C++ seemed
to be the dominant development language. Though programs in C++ are not easily
ported in other operating systems, the benefits in speed, performance and
low-level access of the host system are
invaluable.


\end{document}